\documentclass{article}
\pdfoutput=1
\usepackage{fullpage}
\usepackage{microtype}
\usepackage{graphicx}
\usepackage{subfigure}
\usepackage{booktabs} 
\usepackage{hyperref}
\usepackage{spreadtab}
\usepackage{xspace}
\usepackage{amsfonts, amsmath, amsthm}
\usepackage{pgfplots}
\usepackage{multirow}

\date{}

\newtheorem{proposition}{Proposition}

\usepackage{enumitem}
\setitemize{noitemsep,topsep=0pt,parsep=0pt,partopsep=0pt}

\title{Continuous Integration of Machine Learning Models with ease.ml/ci: Towards a Rigorous Yet Practical Treatment}

\newcommand{\sys}{\texttt{ease.ml/ci}\xspace}

\begin{document}

\author{
{\rm Cedric Renggli}\\
ETH Zurich
\and
{\rm Bojan Karla\v{s}}\\
ETH Zurich
\and 
{\rm Bolin Ding}\\
Alibaba Group
\and 
{\rm Feng Liu}\\
Huawei Technologies
\and 
{\rm Kevin Schawinski}\\
Modulos AG
\and 
{\rm Wentao Wu}\\
Microsoft Research
\and
{\rm Ce Zhang}\\
ETH Zurich
} 

\maketitle


\begin{abstract}
\noindent
Continuous integration is an indispensable step of modern 
software engineering practices to systematically manage
the life cycles of system development. Developing a 
machine learning model is no difference --- it is an 
engineering process with a life cycle, including design, implementation, tuning, testing, and deployment.
However, most, if not all, existing
continuous integration engines do not support
machine learning as first-class citizens.

In this paper, we present \sys, to 
our best knowledge, the first continuous integration system for machine
learning. The challenge of building \sys is to provide
rigorous guarantees, e.g., {\em single accuracy point
error tolerance with 0.999 reliability}, with a practical  
amount of labeling effort, e.g., {\em 2K labels per
test}. We design a domain specific
language that allows users to specify integration conditions 
with reliability constraints, and develop simple novel optimizations
that can lower the number of labels required by up
to two orders of magnitude for test conditions popularly used in real production systems.
\end{abstract}




\section{Introduction}

In modern software engineering~\cite{van2008software}, continuous integration (CI) is an important part of the best practice 
to systematically manage the life cycle of
the development efforts. With a CI engine, 
the practice requires developers to integrate (i.e., commit)
their code into a shared repository at least 
once a day~\cite{duvall2007continuous}. Each commit triggers an automatic build
of the code, followed by running a pre-defined test
suite. The developer receives a \texttt{pass}/\texttt{fail} signal
from each commit, which guarantees that every commit 
that receives a \texttt{pass} signal satisfies
properties that are necessary for product deployment and/or presumed by downstream software.

Developing machine learning models is no different
from developing traditional software, in the sense
that it is also a full life cycle involving
design, implementation, tuning, testing, and deployment.
As machine learning models are used in more task-critical applications and are more tightly integrated with traditional software stacks, 
it becomes increasingly important for the
ML development life cycle also to be managed
following systematic, rigid engineering discipline.
We believe that developing the 
theoretical and system foundation for such a
life cycle management system will be an 
emerging topic for the SysML community.

\begin{figure}
\centering
\includegraphics[width=0.75\textwidth]{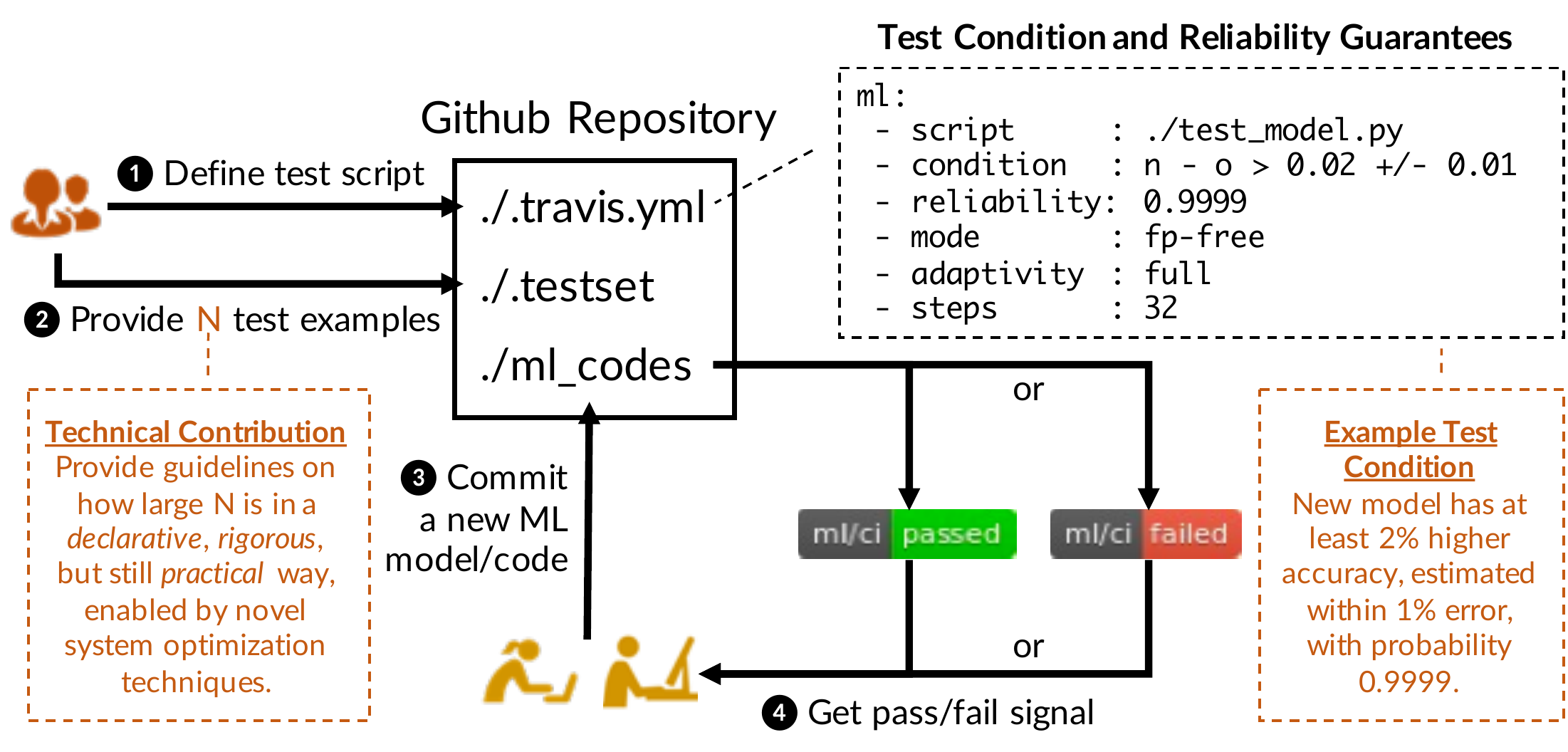}
\caption{The workflow of \sys.}
\label{fig:e2e}
\end{figure}

In this paper, we take the first step towards building,
to our best knowledge, the first continuous integration
system for machine learning. The workflow of the
system largely follows the traditional CI systems
(Figure~\ref{fig:e2e}), while it allows the user
to define machine-learning specific test conditions
such as {\em the new model can only change at most 
10\%
predictions of the old model} or 
{\em the new model must have at least 1\% higher accuracy
than the old model.} After each commit of
a machine learning model/program, the system
automatically tests whether these test conditions
hold, and return a \texttt{pass}/\texttt{fail} signal
to the developer. 
Unlike traditional CI, CI for machine
learning is inherently \emph{probabilistic}. As a result,
all test conditions are evaluated with respect to
a $(\epsilon, \delta)$-reliability requirement
from the user, where $1-\delta$ (e.g., 0.9999) is the probability of a valid test and $\epsilon$ is
the error tolerance (i.e., the length of the
$(1-\delta)$-confidence interval). The
goal of the CI engine is to return 
the \texttt{pass}/\texttt{fail} signal
that satisfies the $(\epsilon, \delta)$-reliability requirement.

\paragraph*{Technical Challenge: Practicality}
At the first glance of the problem, there
seems to exist a trivial implementation:
For each committed model, draw $N$ labeled
data points from the testset, get
an $(\epsilon, \delta)$-estimate of the 
accuracy of the new model, and test whether
it satisfies the test conditions or not.
The challenge of this strategy is 
the practicality associated with the label
complexity (i.e., how large $N$ is). To get
an $(\epsilon = 0.01, \delta = 1- 0.9999)$
estimate of a random variable ranging in $[0, 1]$,
if we simply apply Hoeffding's inequality,
we need more than 46K labels from the user
(similarly, 63K labels for 32 models
in a non-adaptive fashion and 156K
labels in a fully adaptive fashion, see Section~\ref{sec:estimation:basic})!
The technical contribution of this work
is a collection of techniques that 
lower the number of samples, by up to two
orders of magnitude, that the system requires
to achieve the same reliability.

In this paper, we make contributions from both 
the system and machine learning perspectives.

\begin{enumerate}
\item {\bf System Contributions.} We propose
a novel system architecture to support 
a new functionality compensating state-of-the-art ML systems. Specifically, rather than
allowing users to compose adhoc, free-style test conditions, we design a domain specific language that is more restrictive but expressive enough to capture many test conditions of practical interest.
\item {\bf Machine Learning Contributions.}
On the machine learning side, we develop
simple, but novel, optimization techniques 
to optimize for test conditions that can be
expressed within the domain-specific language
that we designed. Our techniques cover 
different modes of interaction (fully 
adaptive, non-adaptive, and hybrid), 
as well as many popular test conditions that industrial and academic partners found useful. 
For {a subset of test conditions}, we 
are able to achieve up to two orders of magnitude savings on the number of labels that the system requires.
\end{enumerate}

Beyond these specific technical contributions,
conceptually, this work illustrates that
enforcing and monitoring an ML development
life cycle in a rigorous way {\em does not need to 
be expensive}. Therefore, ML systems in the near future could afford to 
support more sophisticated monitoring functionality
to enforce the ``right behavior'' from 
the developer.

In the rest of this paper, we start by presenting the design of \sys in Section~\ref{sec:design}. We then develop estimation techniques that can lead to strong probabilistic guarantees using test datasets with moderate labeling effort. We present the basic implementation in Section~\ref{sec:estimation:basic} and more advanced optimizations in Section~\ref{sec:estimation:optimization}.
We further verify the correctness and effectiveness of our estimation techniques via an experimental evaluation (Section~\ref{sec:experiments}).
We discuss related work in Section~\ref{sec:relatedwork} and conclude in Section~\ref{sec:conclusion}.

\section{System Design}
\label{sec:design}

We present the design of \sys in this section. We start by presenting the interaction model and workflow as illustrated in Figure~\ref{fig:e2e}. We then present the scripting language that enables user interactions in a declarative manner. We discuss the syntax and semantics of individual elements, as well as their physical implementations and possible extensions. We end up with two system utilities, a ``sample size estimator'' and a ``new testset alarm,'' the technical details of which will be explained in Sections~\ref{sec:estimation:basic} and~\ref{sec:estimation:optimization}.

\subsection{Interaction Model}

\sys is a {\em continuous integration system}
for machine learning. It supports a four-step
workflow: (1) user describes test conditions in a {\em test configuration script} with respect to the quality of an ML model; (2) user provides $N$ test
examples where $N$ is automatically calculated
by the system given the configuration script; (3) whenever developer commits/checks in 
an updated ML model/program, the
system triggers a build; and (4) the 
system tests whether the test condition is
satisfied and returns a ``pass/fail'' signal
to the developer. When the current testset
loses its ``statistical power'' due to repetitive evaluation, the system also decides
on when to request a new testset from 
the user. The old testset can then be
released to the developer as a validation 
set used for developing new models.

We also distinguish between two teams of people:
the integration team, who provides testset
and sets the reliability requirement; and the 
development team, who commits new models. In practice,
these two teams can be identical; however,
we make this distinction in this paper for clarity,
especially in the fully adaptive case. We call
the integration team {\em the user} and 
the development team {\em the developer}.

\subsection{A \sys Script}

\sys provides a declarative way for users
to specify requirements of a new machine learning model in terms of a set of test cases.
\sys then compiles such specifications into a {\em practical} workflow to enable evaluation of test cases with rigorous theoretical guarantees. 
We present the design of the \sys scripting language, followed by its implementation as an extension to the \texttt{.travis.yml} format used by Travis CI.

\paragraph*{Logical Data Model} The core part of a \sys
script is a user-specified condition for the continuous
integration test. In the current version, such a condition
is specified over three variables $\mathcal{V} = \{n, o, d\}$:
(1) $n$, the accuracy of the new model; (2) $o$, the
accuracy of the old model; and (3) $d$, the \emph{percentage} of new predictions that are different from the old ones ($n, o, d \in [0, 1]$).

A detailed overview over the exact syntax and its semantics is given in Appendix~\ref{app:syn_seman}.

\paragraph*{Adaptive vs. Non-adaptive Integration}
A prominent difference between \sys and 
traditional continuous integration system is that 
the statistical power of a test dataset will decrease
when the result of whether a new model
passes the continuous integration test is released
to the developer. The developer, if she wishes, can adapt her next model to increase its probability to pass the test, as demonstrated by the recent work on adaptive analytics~\cite{blum2015ladder,dwork2015reusable}.
As we will see, ensuring probabilistic guarantee in the adaptive case is more expensive as it requires a larger testset.
\sys allows the user to specify whether the test is adaptive or not with a flag \texttt{adaptivity} (\texttt{full},
\texttt{none}, \texttt{firstChange}):
\begin{itemize}
    \item If the flag is set to \texttt{full}, \sys releases whether the new model passes the test immediately to the developer.
    \item If the flag is set to \texttt{none}, \sys  accepts all commits, however, sends the information of whether the model really passes the test to a user-specified, third-party, email address that
    the developer does not have access to.
    \item If the flag is set to \texttt{firstChange}, \sys allows full adaptivity before the first time that the test passes (or fails), but stops afterwards and requires a new testset (see Section~\ref{sec:estimation:basic} for more details).
\end{itemize}

\paragraph*{Example Scripts} A \sys script is implemented as an 
extension to the \texttt{.travis.yml} file format used in Travis CI by adding an \texttt{ml} section.
For example,
{\small
\begin{verbatim}
  ml:
    - script     : ./test_model.py 
    - condition  : n - o > 0.02 +/- 0.01 
    - reliability: 0.9999
    - mode       : fp-free
    - adaptivity : full
    - steps      : 32
\end{verbatim}
}
This script specifies a continuous test process
that, with probability larger than 0.9999,
accepts the new commit only if the new model 
has two points higher accuracy than the old
one. This estimation is conducted with an 
estimation error within one accuracy point
in a ``false-positive free'' manner.
We give a detailed definition, as well as
a simple example of the two modes \texttt{fp-free} and \texttt{fn-free} in Appendix~\ref{app:semantics}.
The system 
will release the \texttt{pass}/\texttt{fail}
signal immediately to the developer,
and the user expects that the given testset
can be used by as many as 32 times before a new testset has to be provided to the system.

Similarly, if the user wants to specify a 
non-adaptive integration process, she can provide
a script as follows:
{\small
\begin{verbatim}
  ml:
    - script     : ./test_model.py 
    - condition  : d < 0.1 +/- 0.01
    - reliability: 0.9999
    - mode       : fp-free
    - adaptivity : none -> xx@abc.com
    - steps      : 32
\end{verbatim}
}
It accepts each commit but sends
the test result to the email address 
\texttt{xx@abc.com} after each commit.
The assumption is that the developer does
not have access to this email account and
therefore, cannot adapt her next model.

\paragraph*{Discussion and Future Extensions}
The current syntax of \sys is able to capture 
many use cases that our users find
useful in their own development process, including to reason about the accuracy difference between the new and old models,
and to reason about the amount of changes in predictions between the new and old models in the test dataset. In principle,
\sys can support a richer syntax.
We list some limitations of the current
syntax that we believe are interesting directions
for future work.
\begin{enumerate}
\item Beyond accuracy: There are other important
quality metrics for machine learning that the current
system does not support, e.g., F1-score, AUC score, etc.
It is possible to extend the current system to accommodate these
scores by replacing the Bennett's inequality 
with the McDiarmid's inequality, together with the sensitivity of
F1-score and AUC score. In this new context, more optimizations,
such as using stratified samples, are possible for skewed cases.
\item Ratio statistics: The current syntax of \sys
intentionally leaves out division (``\texttt{/}'') and it
would be useful for a future version to enable \emph{relative} comparison of qualities (e.g., accuracy, F1-score, etc.).
\item Order statistics: Some users
think that order statistics are also useful, e.g.,
to make sure the new model is among top-5 models in the development history.
\end{enumerate}
Another limitation of the current system is the lack of being able to detect a domain drift or concept ship. In principle, this could be thought of as a similar process of CI -- instead of fixing the test set and testing multiple models, monitoring concept shift is to fix a single model and test its generalization over multiple test sets overtime.

The current version of \sys does not provide support
for all these features. However, we believe that many
of them can be supported by developing similar statistical
techniques (see Sections~\ref{sec:estimation:basic} and~\ref{sec:estimation:optimization}).

\subsection{System Utilities}
\label{sec:design:utilities}

In traditional continuous integration, the system often assumes that the user has the knowledge and competency to build the test suite all by herself.
This assumption is too strong for \sys --- among
the current users of \sys, we observe that even
experienced software engineers in large 
tech companies can be clueless on how to 
develop a proper testset for a given reliability requirement.
One prominent contribution of \sys is a collection of techniques
that provide practical, but rigorous, guidelines 
for the user to manage testsets: 
{\em How large
does the testset need to be?}
{\em When does the system need
a new freshly generated testset?}
{\em When can
the system release the testset and ``downgrade''
it into a development set?}
While most of these
questions can be answered by experts based on heuristics
and intuition, the goal of \sys is to provide 
systematic, principled guidelines.
To achieve this goal, \sys provides two utilities 
that are not provided in systems such as Travis CI.

\paragraph*{Sample Size Estimator} This is a program that takes as input a \sys script, and outputs the number of examples that the user needs to provide in the testset. 

\paragraph*{New Testset Alarm}
This subsystem is a program
that takes as input a \sys script as well as the commit history of machine learning models,
and produces an alarm (e.g., by 
sending an email) to the user when the current
testset has been used too many times and thus cannot
be used to test the next committed model.
Upon receiving the alarm, the user needs to provide 
a new testset to the system and can also release the old
testset to the developer.

An impractical implementation of these two utilities
is easy --- the system alarms the user to request
a new testset after every commit and estimates the testset size using the Hoeffding bound.
However, this can result in testsets that require tremendous labeling effort, which is not always feasible.

\paragraph*{What is ``Practical?''} The practicality 
is certainly user dependent. Nonetheless, from our experience working
with different users, we observe that 
providing $30,000$ to $60,000$ labels for every $32$ model
evaluations seems reasonable for many users: 
$30,000$ to $60,000$ is what 2 to 4
engineers can label in a day (8 hours) at a
rate of 2 seconds per label, and 32 model evaluations
imply (on average) one commit per day in a month. Under this assumption, the user only needs to spend one day per month to provide test labels with a reasonable number of labelers. If the user is not able to provide this amount of labels, a ``cheap mode'', where the number of labels per day is easily reduced by a factor 10x, is achieved for most of the common conditions by increasing the error tolerance by a single or two percentage points.

Therefore, to make \sys a useful tool for real-world 
users, these utilities need to be implemented in 
a more practical way. The technical contribution of
\sys is a set of techniques that we will present next, which can reduce 
the number of samples the system requests from the user
by up to two orders of magnitude.

\section{Baseline Implementation}
\label{sec:estimation:basic}

We describe the techniques to implement 
\sys for user-specified conditions in the
most general case. The techniques that we
use involve standard Hoeffding inequality
and a technique similar to Ladder~\cite{blum2015ladder}
in the adaptive case. This implementation is
general enough to support all user-specified conditions
currently supported in \sys, however, it can
be made more practical when the test conditions
satisfy certain conditions. We leave optimizations for specific conditions to Section~\ref{sec:estimation:optimization}.

\subsection{Sample Size Estimator for a Single Model}

\paragraph*{Estimator for a Single Variable}
One building block of \sys is the estimator of
the number of samples one needs to estimate 
one variable ($n$, $o$, and $d$) to $\epsilon$
accuracy with $1-\delta$ probability. We construct
this estimator using the standard Hoeffding bound.

A sample size estimator $n: \mathcal{V} \times [0,1]^3 \mapsto \mathbb{N}$ is a function that takes as input
a variable, its dynamic range, error tolerance and
success rate, and outputs the number of samples one needs in a testset. With the standard Hoeffding bound,
\[
n(v, r_v, \epsilon, \delta) = \frac{-r_v^2 \ln \delta}{2\epsilon^2}
\]
where $r_v$ is the dynamic range of the variable 
$v$, $\epsilon$ the error tolerance, and 
$1-\delta$ the success probability.

Recall that we makes use of the exact grammar 
used to define the test conditions. A formal definition
of the syntax can be found in Appendix~\ref{app:syntax}.

\paragraph*{Estimator for a Single Clause}

Given a clause $C$ (e.g. $n - o > 0.01$) with a left-hand side expression
$\Phi$, a comparison operator \texttt{cmp} ($>$ or $<$), and a right-hand
side constant, the sample size estimator returns the
number of samples one needs to provide an 
$(\epsilon, \delta)$-estimation of the left-hand side 
expression. This can be done with a 
trivial recursion:
\begin{enumerate}
\item $n(\texttt{EXP} = \texttt{c * v}, \epsilon, \delta) = n(v, r_v, \epsilon / c, \delta)$, where $c$ is a constant. We have
$n(\texttt{c * v}, \epsilon, \delta) = \frac{- c^2 r_v^2 \ln \delta}{2\epsilon^2}$.
\item $n(\texttt{EXP1 + EXP2}, \epsilon, \delta) =
   \max \{n(\texttt{EXP1}, \epsilon_1, \frac{\delta}{2}),$
          $n(\texttt{EXP2}, \epsilon_2, \frac{\delta}{2})\}$,
where $\epsilon_1 + \epsilon_2 < \epsilon$. The same
equality holds similarly for
$n(\texttt{EXP1 - EXP2}, \epsilon, \delta)$.
\end{enumerate}

\paragraph*{Estimator for a Single Formula}

Given a formula $F$ that is a conjunction over
$k$ clauses ${C_1, ..., C_k}$, the sample size
estimator needs to guarantee that it can 
satisfy each of the clause $C_i$. One way
to build such an estimator is
\begin{enumerate}
\setcounter{enumi}{2}
\item $n(\texttt{F} = C_1 \wedge \ldots \wedge C_k, \epsilon, \delta) = \max_i n(C_i, \epsilon, \frac{\delta}{k})$.
\end{enumerate}

\paragraph*{Example} Given a formula $F$, we now have a simple algorithm for sample size estimation. 
For
{\small
\begin{verbatim}
 F :- n - 1.1 * o > 0.01 +/- 0.01 /\ d < 0.1 +/- 0.01    
\end{verbatim}
}
the system solves an
optimization problem:
\[
n(\texttt{F}, \epsilon, \delta) = 
\min_{\substack{
\epsilon_1 + \epsilon_2 = \epsilon\\ \epsilon_1, \epsilon_2 \in [0, 1]}}
\max\{
\frac{- \ln \frac{\delta}{4}}{2\epsilon_1^2},
\frac{- 1.1^2 \ln \frac{\delta}{4}}{2\epsilon_2^2},
\frac{- \ln \frac{\delta}{2}}{2\epsilon^2}
\}.
\]

\subsection{Non-Adaptive Scenarios}

In the non-adaptive scenario, the system evaluates 
$H$ models, without releasing the result to 
the developer. The result can be released to 
the user (the integration team).

\paragraph*{Sample Size Estimation} Estimation of sample
size is easy in this case because all $H$ models are independent.
With probability $1-\delta$, \sys returns the right answer for each of the
$H$ models, the number of samples one needs for formula $F$ is simply
$n(\texttt{F}, \epsilon, \frac{\delta}{H}).$
This follows from the standard union bound. 
Given the number of models that
user hopes to evaluate (specified 
in the \texttt{steps} field of a \sys script),
the system can then return the number of samples in the testset.

\paragraph*{New Testset Alarm} The alarm 
for users to provide a new testset is
easy to implement in the non-adaptive scenario.
The system maintains a counter of how many 
times the testset has been used. When 
this counter reaches the pre-defined 
budget (i.e., \texttt{steps}), the system
requests a new testset from the user.
In the meantime, the old testset can be 
released to the developer for future development
process.

\subsection{Fully-Adaptive Scenarios}

In the fully-adaptive scenario, the system releases
the test result (a single bit indicating pass/fail)
to the developer. Because this bit leaks information
from the testset to the developer, one cannot use
union bound anymore as in the non-adaptive scenario.

A trivial strategy exists for such a case --- for
every model, uses a different testset. In this case,
the number of samples required is 
$H \cdot n(\texttt{F}, \epsilon, \frac{\delta}{H})$.
This can be improved by applying a adaptive argument similar to Ladder~\cite{blum2015ladder} as follows.

\paragraph*{Sample Size Estimation} For the 
fully adaptive scenario, \sys uses the following way
to estimate the sample size for an $H$-step 
process. The intuition is simple.
Assume that a developer is deterministic 
or pseudo-random, her decision on the next model only
relies on all the previous \texttt{pass}/\texttt{fail} signals and the initial model $H_0$. 
For $H$ steps, there are
only $2^H$ possible configurations of
the past \texttt{pass}/\texttt{fail} signals. As a result, one
only needs to enforce the union bound on all
these $2^H$ possibilities. Therefore, the number
of samples one needs is
$n(\texttt{F}, \epsilon, \frac{\delta}{2^H}).$

\paragraph*{Is the Exponential Term too Impractical?}
The improved sample size $n(\texttt{F}, \epsilon, \frac{\delta}{2^H})$ is much smaller than the one, $H \cdot n(\texttt{F}, \epsilon, \frac{\delta}{H})$, required by the trivial strategy.
Readers might worry about the dependency 
on $H$ for the fully adaptive scenario. However, 
for $H$ that is not too large, e.g., $H=32$, 
the above bound can still lead to practical 
number of samples as the $\frac{\delta}{2^H}$ is
within a logarithm term. As an example, consider the
following simple condition:

{\small
\begin{verbatim}
          F :- n > 0.8 +/- 0.05.
\end{verbatim}
}
With $H=32$, we have
\[
n(\texttt{F}, \epsilon, \frac{\delta}{2^H}) = \frac{\ln 2^H - \ln \delta}{2\epsilon^2}.
\]
Take $\delta = 0.0001$ and $\epsilon = 0.05$, 
we have $n(\texttt{F}, \epsilon, \frac{\delta}{2^H}) = 6,279$. Assuming the developer checks in the best model
everyday, this means that every month the user needs
to provide only fewer than seven thousand test samples,
a requirement that is not too crazy.
However, if $\epsilon = 0.01$, this blows up to $156,955$, which
is less practical. We will show how to tighten this bound in Section~\ref{sec:estimation:optimization} for a sub-family of test conditions.

\paragraph*{New Testset Alarm} Similar to the non-adaptive
scenario, the alarm for requesting a new testset is 
trivial to implement --- the system requests a new
testset when it reaches the pre-defined budget. At that
point, the system can release the testset to the developer
for future development.

\subsection{Hybrid Scenarios}

One can obtain a better bound on the number of required
samples by constraining the information being released
to the developer. Consider the following scenario:
\begin{enumerate}
\item If a commit fails, returns \texttt{Fail} to the developer;
\item If a commit passes, (1) returns \texttt{Pass} to the developer, and (2) triggers the new testset alarm to request a new testset from the user.
\end{enumerate}
Compared with the fully adaptive scenario, in this
scenario, the user provides a new testset immediately
after the developer commits a model that passes the 
test. 

\paragraph*{Sample Size Estimation}
Let $H$ be the maximum number of steps the system supports.
Because the system will request a new testset
immediately after a model passes the test, it is not really adaptive: 
As long as the developer continues to use the same testset, she
can assume that the last model always fails. Assume
that the user is a deterministic function that
returns a new model given the past history and 
past feedback (a stream of \texttt{Fail}), there are
only $H$ possible states that we need to apply
union bound. This gives us the same bound as 
the non-adaptive scenario:
$n(\texttt{F}, \epsilon, \frac{\delta}{H})$.

\paragraph*{New Testset Alarm} Unlike the
previous two scenarios, the system will alarm
the user whenever the model that she provides passes
the test or reaches the pre-defined budget $H$, whichever comes earlier.

\paragraph*{Discussion} It might be counter-intuitive that the hybrid scenario, which leaks information to the developer, has the
same sample size estimator as the non-adaptive case.
Given the maximum number of steps that the testset
supports, $H$, the hybrid scenario cannot always 
finish all $H$ steps as it might require a new testset
in $H' \ll H$ steps. In other words, in contrast to the fully adaptive scenario, the hybrid scenario accommodates the leaking of information not by adding more samples, but by decreasing the number of steps that a
testset can support. 

The hybrid scenario is useful when the test is hard to pass or fail.
For example, imagine the following condition: 
{\small
\begin{verbatim}
        F :- n - o > 0.1 +/- 0.01
\end{verbatim}
}
That is, the system only accepts commits that increase the
accuracy by 10 accuracy points. In this case, the developer
might take many developing iterations to get a model
that actually satisfies the condition. 

\subsection{Evaluation of a Condition}

Given a testset that satisfies the number of
samples given by the sample size estimator,
we obtain the estimates of the three variables used in a clause, i.e.,
$\hat{n}$, $\hat{o}$, and $\hat{d}$. Simply
using these estimates to evaluate a condition
might cause both false positives and false negatives.
In \sys, we instead replace the point estimates by their corresponding confidence intervals, and define a simple algebra over
intervals (e.g., $[a,b]+[c,d]=[a+c, b+d]$), which is used to evaluate the left-hand side of a single clause. A clause still evaluates to \{\texttt{True}, \texttt{False}, \texttt{Unknown}\}. The system then maps
this three-value logic into a two-value logic
given user's choice of either \texttt{fp-free}
or \texttt{fn-free}. 

\subsection{Use Cases and Practicality Analysis}

The baseline implementation of \sys relies on 
standard concentration bounds with simple,
but novel, twists to the specific use cases. 
Despite its simplicity, this implementation
can support real-world scenarios 
that many of our users find useful. We summarize 
five use cases and analyze the number of samples required from the user. 
These use cases are summarized from observing the 
requirements from the set of users we have been 
supporting over the last two years, ranging from
scientists at multiple universities, to real production 
applications provided by high-tech companies. 
($\texttt{[c]}$ and $\texttt{[epsilon]}$ are placeholders for constants.)

\vspace{1em}
\noindent 
{\bf (F1: Lower Bound Worst Case Quality)}
{\small
\begin{verbatim}
  F1         :- n > [c] +/- [epsilon]
  adaptivity :- none
  mode       :- fn-free
\end{verbatim}
}
This condition is used for quality control to avoid
the cases that the developer accidentally commits a
model that has an unacceptably low quality or
has obvious quality bugs.
We see many use cases of this condition in non-adaptive scenario, most of which need to be false-negative free.


\vspace{3em}
\noindent 
{\bf (F2: Incremental Quality Improvement)}
{\small
\begin{verbatim}
  F2         :- n - o > [c] +/- [epsilon]
  adaptivity :- full
  mode       :- fp-free
  ([c] is small)
\end{verbatim}
}
This condition is used for making sure that the
machine learning application monotonically improves
over time. This is important when the machine
learning application is end-user facing, in which 
it is unacceptable for the quality to drop.
In this scenario, it makes sense for the whole
process to be fully adaptive and false-positive free.

\vspace{1em}
\noindent 
{\bf (F3: Significant Quality Milestones)}
{\small
\begin{verbatim}
  F3         :- n - o > [c] +/- [epsilon]
  adaptivity :- firstChange
  mode       :- fp-free
  ([c] is large)
\end{verbatim}
}
This condition is used for making sure that the
repository only contains significant quality
milestones (e.g., log models after 10 points
of accuracy jump). Although the condition is
syntactically the same as \texttt{F2}, it makes sense for the whole
process to be hybrid adaptive and false-positive free.

\vspace{1em}
\noindent 
{\bf (F4: No Significant Changes)}
{\small
\begin{verbatim}
  F4         :- d < [c] +/- [epsilon]
  adaptivity :- full | none
  mode       :- fn-free
  ([c] is large)
\end{verbatim}
}
This condition is used for safety concerns 
similar to \texttt{F1}. When the machine
learning application is end-user facing or
part of a larger application, it is important
that its prediction will not change significantly
between two subsequent versions.
Here, the process needs to be false-negative
free. Meanwhile, we see use cases for both
fully adaptive and non-adative scenarios.

\vspace{1em}
\noindent 
{\bf (F5: Compositional Conditions)}
{\small
\begin{verbatim}
   F5 :- F4 /\ F2
\end{verbatim}
}
One of the most popular test conditions
is a conjunction of two conditions,
\texttt{F4} and \texttt{F2}: 
The integration team wants to 
use \texttt{F4} and \texttt{F2} together so that
the end-user facing application will not experience dramatic quality change.


\paragraph*{Practicality Analysis} How practical
is it for our baseline implementation to support
these conditions, and in which case that 
the baseline implementation becomes impractical?

\begin{figure}
%
%
\scriptsize
\centering
\begin{tabular}{ll|rr|rr}
\hline
\multirow{2}{*}{1-$\delta$} &
\multirow{2}{*}{ $\epsilon$} & 
\multicolumn{2}{c|}{\texttt{F1}, \texttt{F4}} &
\multicolumn{2}{c}{\texttt{F2}, \texttt{F3}} \\
 & & \texttt{none} & \texttt{full} &  
 \texttt{none} & \texttt{full}  \\
\hline
0.99 & 0.1 & 404 & 1340 & 1753 & 5496 \\
0.99 & 0.05 & 1615 & 5358 & 7012 & 21984 \\
0.99 & 0.025 & 6457 & 21429 & 28045 & \textcolor{red}{87933} \\
0.99 & 0.01 & 40355 & \textcolor{red}{133930} & \textcolor{red}{175282} & \textcolor{red}{549581} \\
\hline
0.999 & 0.1 & 519 & 1455 & 2214 & 5957 \\
0.999 & 0.05 & 2075 & 5818 & 8854 & 23826 \\
0.999 & 0.025 & 8299 & 23271 & 35414 & \textcolor{red}{95302} \\
0.999 & 0.01 & 51868 & \textcolor{red}{145443} & \textcolor{red}{221333} & \textcolor{red}{595633} \\
\hline
0.9999 & 0.1 & 634 & 1570 & 2674 & 6417 \\
0.9999 & 0.05 & 2536 & 6279 & 10696 & 25668 \\
0.9999 & 0.025 & 10141 & 25113 & 42782 & \textcolor{red}{102670} \\
0.9999 & 0.01 & \textcolor{red}{63381} & \textcolor{red}{156956} & \textcolor{red}{267385} & \textcolor{red}{641684} \\
\hline
0.99999 & 0.1 & 749 & 1685 & 3135 & 6878 \\
0.99999 & 0.05 & 2996 & 6739 & 12538 & 27510 \\
0.99999 & 0.025 & 11983 & 26955 & 50150 & \textcolor{red}{110038} \\
0.99999 & 0.01 & \textcolor{red}{74894} & \textcolor{red}{168469} & \textcolor{red}{313437} & \textcolor{red}{687736} \\
\hline
\end{tabular}
\caption{Number of samples required by different conditions, $H=32$ steps. Red font indicates ``impractical'' number of samples (see discussion on practicality in Section~\ref{sec:design:utilities}).}
\vskip -4ex
\label{fig:naive}
\end{figure}

\paragraph*{When is the Baseline Implementation Practical?}
The baseline implementation, in spite of its simplicity, is practical in many cases. Figure~\ref{fig:naive}
illustrates the number of samples the system requires
for $H=32$ steps. We see that, for both \texttt{F1}
and \texttt{F4}, all adaptive strategies are 
practical up to 2.5 accuracy points, while for
\texttt{F2} and \texttt{F3}, the non-adaptive and
hybrid adaptive strategies are practical up to 2.5
accuracy points 
and the fully adaptive strategy
is only practical up to 5 accuracy points.
As we see from this example, even with a simple 
implementation, {\em enforcing a rigorous guarantee
for CI of machine learning 
is not always expensive!}

\paragraph*{When is the Baseline Implementation Not Practical?}
We can see from Figure~\ref{fig:naive} the strong dependency
on $\epsilon$. This is expected because of the
$O(1/\epsilon^2)$ term in the Hoeffding inequality.
As a result, none of the adaptive strategy is
practical up to 1 accuracy point, a level of 
tolerance that is important for many task-critical 
applications of machine learning. It is also not 
surprising that the fully adaptive strategy requires
more samples than the non-adaptive one, and therefore
becomes impractical with higher error tolerance.

\section{Optimizations}
\label{sec:estimation:optimization}

As we see from the previous sections, the baseline
implementation of \sys fails to provide a practical
approach for low error tolerance and/or fully adaptive
cases. In this section, we describe optimizations
that allow us to further improve the sample size estimator.

\paragraph*{High-level Intuition} All of
our proposed techniques in this section are based on the
same intuition: Tightening the sample size
estimator in the worst case is hard 
to get better than $O(1/\epsilon^2)$;
instead, we take 
the classic system way of thinking --- {\em improve
the the sample size estimator for a 
sub-family of popular test conditions}.
Accordingly, \sys applies different optimization techniques for test conditions of different forms.

\paragraph*{Technical Observation 1} The 
intuition behind a tighter sample
size estimator relies on standard 
techniques of tightening Hoeffding's
inequality for variables with 
small variance. Specifically, when
the new
model and the old model is only
different on up to $(100\times p )\%$
of the predictions, which could be 
part of the test condition anyway, for data point
$i$, the random variable $n_i - o_i$
has small variance:
$\mathbb{E}\left[(n_i - o_i)^2\right] < p$,
where $n_i$ and $o_i$ are the 
predictions of the new and old models on the data point $i$.
This allows us to apply the standard Bennett's inequality.

\begin{proposition}[Bennett's inequality]
Let $X_1, ..., X_n$ be
independent and square integrable random variables such that for some nonnegative constant $b$, $|X_i| \leq b$
almost surely for all $i < n$. We have
\[
\Pr\left[ \left| \frac{\sum_i X_i - \mathbb{E}[X_i]}{n} \right| > \epsilon \right] \le
2\exp\left( -\frac{v}{b^2} h\left(\frac{nb\epsilon}{v}\right) \right),
\]
where $v = \sum_i \mathbb{E}\left[X_i^2\right]$ and
$h(u) = (1+u)\ln(1+u) - u$
for all positive $u$.
\end{proposition}

\paragraph*{Technical Observation 2} The second
technical observation is that, 
to estimate 
the difference of predictions between the new
model and the old model, one does not need
to have labels. Instead, a sample from
the unlabeled dataset is enough to estimate 
the difference. Moreover, to estimate 
$n - o$ when only $10\%$ data points 
have different predictions, one only needs to
provide labels to $10\%$ of the whole testset.

\subsection{Pattern 1: Difference-based Optimization}

The first pattern that \sys searches in a
formula is whether it is of the following form
{\small
\begin{verbatim}
   d < A +/- B /\ n - o > C +/- D 
\end{verbatim}
}
which constrains the amount of changes that
a new model is allowed to have while ensuring
that the new model is no worse than the old model.
These two clauses popularly appear in test
conditions from our users: For production-level
systems, developers start from an already 
good enough, deployed model, and spend most of their time {\em fine-tuning} a machine
learning model. As a result, the continuous 
integration test must have an error tolerance as low as a single accuracy point.
On the other hand, the new model will not be different from the old model significantly, otherwise more engaged
debugging and investigations are 
almost inevitable.

\paragraph*{Assumption.} One assumption of this 
optimization is that it is relatively cheap
to obtain unlabeled data samples, whereas it is 
expensive to provide labels. This is true in many
of the applications. When this assumption is
valid, both optimizations in 
Section~\ref{sec:op1.1} and Section~\ref{sec:op1.2}
can be applied to this pattern; otherwise,
both optimizations still apply but will lead to improvement over only a subset.

\subsubsection{Hierarchical Testing} \label{sec:op1.1}

The first optimization is to test the rest of the
clauses conditioned on 
\texttt{d < A +/- B}, which
leads to an algorithm with two-level tests. The first level tests
whether the difference between the
new model and the old model is small
enough, whereas the second level tests $(n-o)$.

The algorithm runs in two steps:
\begin{enumerate}
\item {\bf (Filter)} Get an $(\epsilon', \frac{\delta}{2})$-estimator $\hat{d}$ with $n'$ samples.
Test whether $\hat{d} > A + \epsilon'$: If 
so, returns \texttt{False};
\item {\bf (Test)} Test \texttt{F} as in the baseline
implementation (with $1 - \frac{\delta}{2}$ probability), conditioned on 
$d < A + 2\epsilon'$.
\end{enumerate}

It is not hard to see why the above algorithm works --- the first step only requires
unlabeled data points and does not need
human intervention.
In the second step, conditioned on $d < p$, we know that
$
\mathbb{E}\left[(n_i - o_i)^2\right] < p
$
for each data point. Combined with 
$|n_i - o_i| < 1$, applying Bennett's inequality we have
$
\Pr \left[ \left| \widehat{n - o} - (n-o) \right| > \epsilon \right]
\leq 2 \exp(-np h\left(\frac{\epsilon}{p}\right) ).
$

As a result, the second step needs a sample size 
(for non-adaptive scenario) of
\[
n = \frac{\ln H -\ln \frac{\delta}{4}}{p h\left(\frac{\epsilon}{p}\right)}.
\]
When $p=0.1, 1 - \delta=0.9999, d < 0.1$, we
only need 29K samples for 32 non-adaptive
steps and 67K samples for 32 fully-adaptive
steps to reach an error tolerance of a single accuracy point --- 10$\times$ fewer than the baseline (Figure~\ref{fig:naive}).

\subsubsection{Active Labeling} \label{sec:op1.2}

The previous example gives the user a way
to conduct 32 fully-adaptive 
fine-tuning steps with only 67K samples.
Assume that the developer performs one commit per
day, this means that we require 67K samples
per month to support the continuous
integration service.

One potential challenge for this strategy
is that all 67K samples need to be labeled
before the continuous integration service
can start working. This is sometimes a 
strong assumption that many users find problematic. 
In the ideal case, we hope to interleave the development effort
with the labeling effort, and amortize 
the labeling effort over time.

The second technique our 
system uses relies on the observation that,
to estimate $(n-o)$, only the data points
that have a different prediction
between the new and old models need to
be labeled. When we know that the
new model predictions are only different from the
old model by $10\%$, we only need
to label $10\%$ of all data points.
It is easy to see that, every time when the developer commits a new model,
we only need to provide
\[
n = \frac{-\ln \frac{\delta}{4}}{p h\left(\frac{\epsilon}{p}\right)} \times p
\]
labels. When $p=0.1$ and $1 - \delta=0.9999$, then
$n = 2188$ for an error tolerance of a single accuracy point.
If the developer commits
one model per day, the labeling 
team only needs to label 2,188 samples
the next day. Given a 
well designed interface that enables
a labeling throughput of 5 seconds
per label, the labeling team only needs
to commit 3 hours a day! For a team with
multiple engineers, this overhead is often acceptable, considering the guarantee provided by the system down to a single accuracy point.

Notice that active labeling assumes a stationary underlying distribution. One way to enforce this in the system is to ask the user to provide a pool of unlabeled data points at the same time, and then only ask for labels when needed. In this way, we do not need to draw new samples over time.

\subsection{Pattern 2: Implicit Variance Bound}

In many cases, the user does not provide an
explicit constraint on the difference between
a new model and an old model. However,
many machine learning models are not
so different in their predictions.
Take AlexNet, ResNet, GoogLeNet, AlexNet (Batch Normalized), and VGG for example: When applied to the ImageNet testset, these
five models, developed by the ML community since 2012, only produce up to 
25\% different answers for top-1 {\em correctness} and
15\% different answers for top-5 {\em correctness}!
For a typical workload of continuous integration,
it is therefore not unreasonable to expect many of the
consecutive commits would have smaller difference
than these ImageNet winners involving
years of development.

Motivated by this observation, \sys will automatically match
with the following pattern
{\small
\begin{verbatim}
             n - o > C +/- D. 
\end{verbatim}
}
When the unlabeled testset is cheap to get,
the system will use one testset
to estimate $d$ up to $\epsilon=2D$: For binary classification task,
the system can use an unlabeled testset;
for multi-class tasks, one can either
test the {\em difference of predictions}
on an unlabeled testset or {\em difference 
of correctness} on a labeled testset.
This gives us an upper bound
of $n - o$. The system then tests 
$n - o$ up to $\epsilon=D$ on  
{\em another} testset (different from
the one used to test $d$).
When this upper bound is
small enough, the
system will trigger similar optimization
as in \texttt{Pattern 1}.
Note that the first testset will be 
16$\times$ smaller than testing
$n - o$ directly up to $\epsilon=D$ --- 4$\times$ due to
a higher error tolerance, and 
4$\times$ due to that $d$
has 2$\times$ smaller range than
$n - o$.

One caveat of this approach is that the system
does not know how large the second
testset would be before execution. The system uses a technique similar to active labeling by incrementally
growing the labeled testset every time when a new model is committed,
if necessary. Specifically, we optimize for test conditions following the pattern
{\small
\begin{verbatim}
                 n > A +/- B, 
\end{verbatim}
}
when \texttt{A} is large (e.g., 0.9 or 0.95).
This can be done by first having a coarse estimation
of the lower bound of $n$, and then conducting 
a finer-grained estimation conditioned on this lower
bound. Note that this can only introduce improvement
when the lower bound is large (e.g., 0.9).

\subsection{Tight Numerical Bounds}

Following~\cite{langford2005tutorial}, having a test condition 
consisting of $n$ i.i.d random variables drawn from a Bernoulli distribution,
one can simply derive a tight bound on the number of samples
required to reach a ($\epsilon$, $\delta$) accuracy. The calculation of number of samples require the probability mass function of the Binomial distribution (sum of i.i.d Bernoulli variables). Tight bound are solved by taking the minimum of number of samples $n$ needed, over the max unknown true mean $p$. This technique can also be extended to more complex queries, where the binomial
distribution has to be replaced by a multimodal distribution. The exact analysis
has, as for the simple case, no closed-form solution, and deriving efficient approximations is left as further work.



\clearpage
\section{Experiments}
\label{sec:experiments}

We focus on empirically validating the derived bounds and show \sys in 
action next. 

\subsection{Sample Size Estimator}

One key technique most of our optimizations relied on
is that, by knowing an upper bound of the sample variance,
we are able to achieve a tighter bound than simply
applying the Hoeffding bound. This upper bound can either be
achieved by using unlabeled data points to estimate
the difference between the new and old models, or 
by using labeled data points but conducting a
coarse estimation first. We now validate our theoretical bound and its impact on improving the label complexity.

\begin{figure}
\centering
\includegraphics[width=0.75\textwidth]{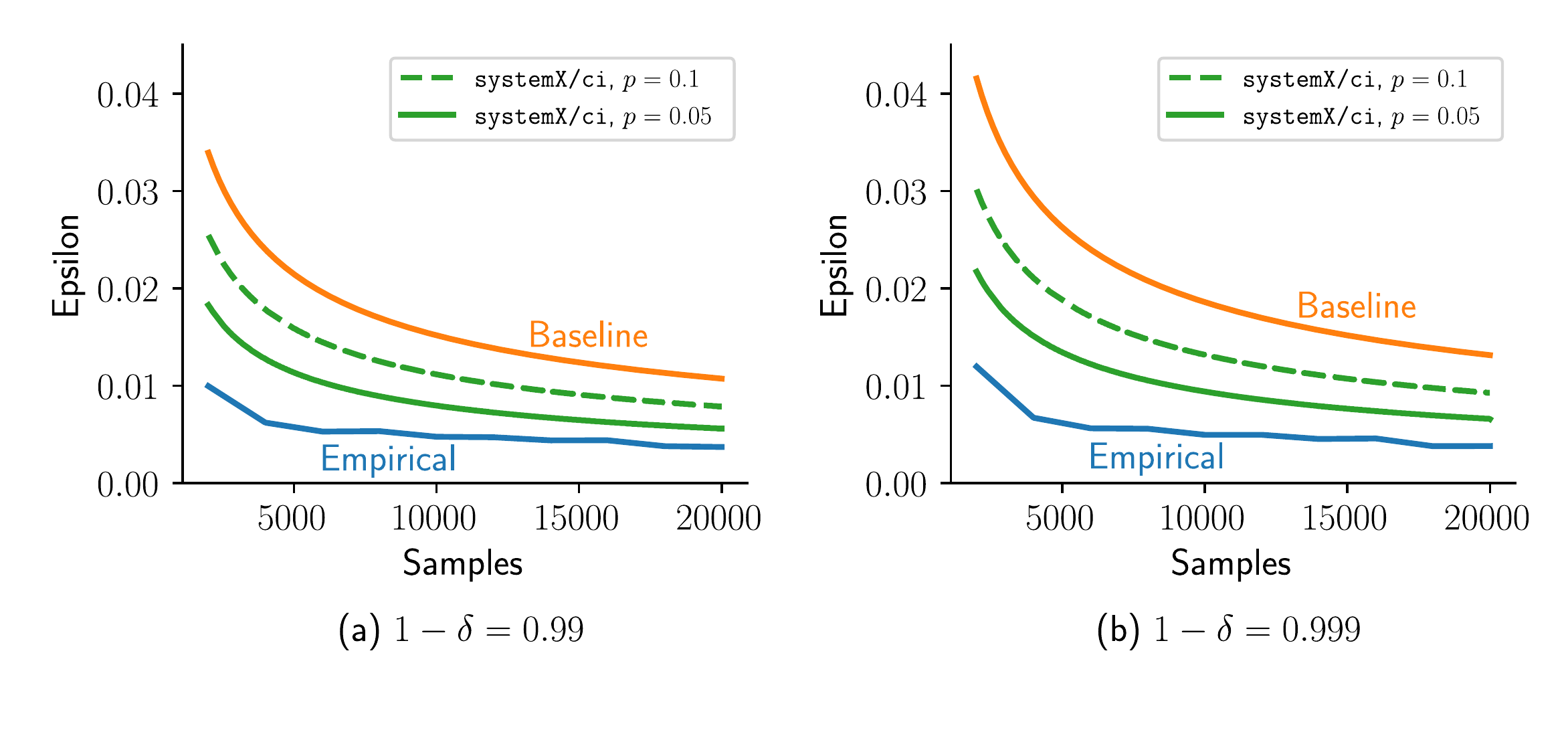}
\caption{Comparison of Sample Size Estimators in
the Baseline Implementation and the Optimized Implementation.}
\label{fig:empirical}
\end{figure}

Figure~\ref{fig:empirical} illustrates the estimated
error and the empirical error by {\em assuming} different
upper bounds $p$, for a model with accuracy around 
98\%. We run GoogLeNet~\cite{jia2014caffe} on the infinite MNIST dataset~\cite{infiniteMNIST} and estimate the true accuracy $c$.
Assuming a non-adaptive scenario, we obtain a range of accuracies achieved by randomly taking $n$ data points. We then estimate the interval $\epsilon$ 
with the given number of samples $n$ and probability $1-\delta$. We see that, both the baseline
implementation and \sys dominate the empirical 
error, as expected, while \sys uses significantly
fewer samples.\footnote{The empirical error was determined by taking different testsets (with the sample sample size) and measuring the gap between the $\delta$ and $1-\delta$ quantiles over the observed testing accuracies.}

\begin{figure}
\centering
\includegraphics[width=0.75\textwidth]{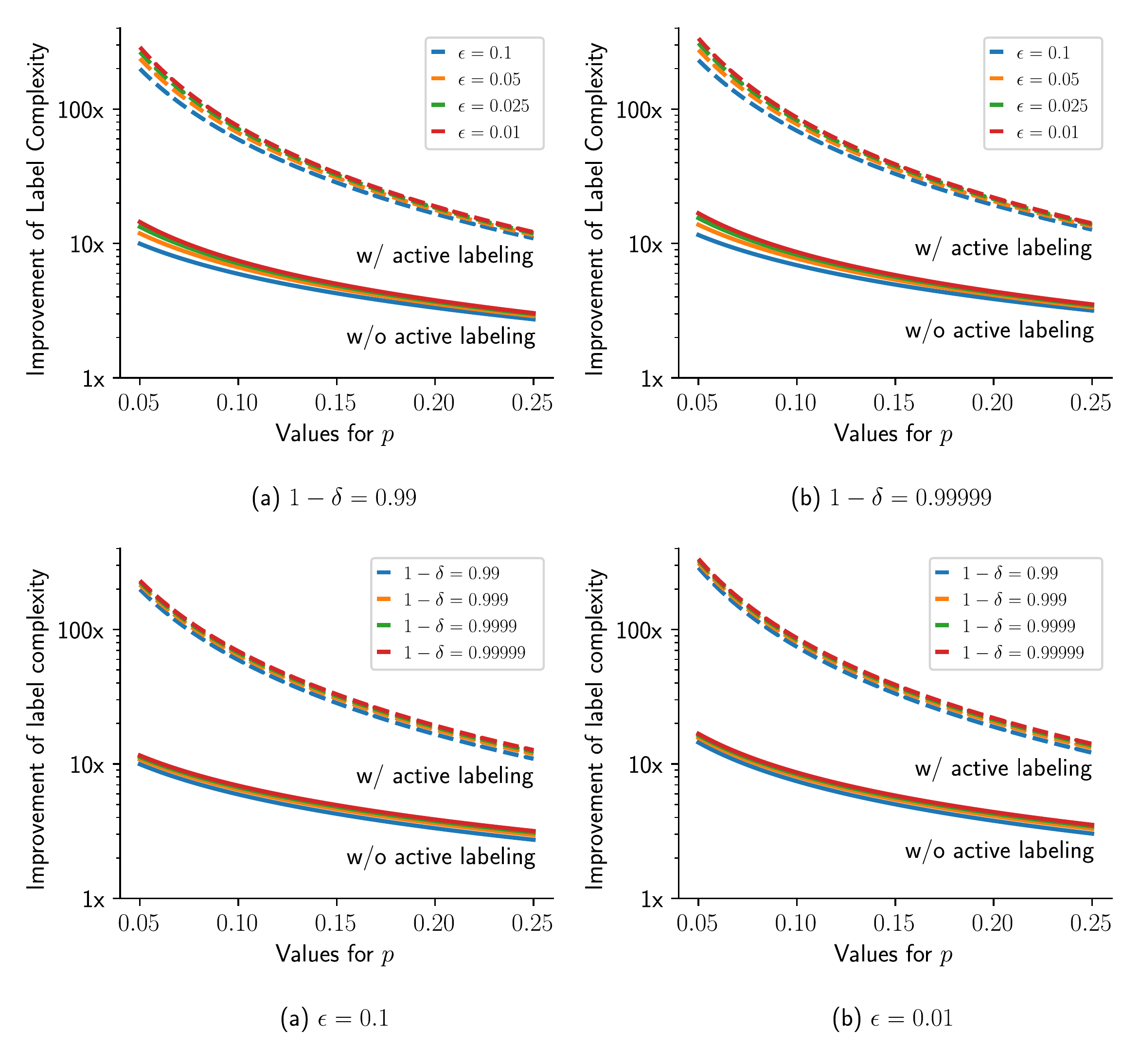}
\caption{Impact of $\epsilon$, $\delta$, and $p$ on the Label Complexity.}
\label{fig:savings}
\end{figure}

Figure~\ref{fig:savings} illustrates the impact
of this upper bound on improving the label complexity.
We see that, the improvement increases significantly when $p$ is reasonably 
small --- when $p = 0.1$, we can achieve almost
$10\times$ improvement on the label complexity.
Active labeling further increases the improvement,
as expected, by another 10$\times$.

\subsection{\sys in Action}

We showcase three different test conditions for a real-world incremental development of machine learning models submitted to the SemEval-2019 Task 3 competition. The goal is to classify the emotion of the user utterance as one of the following classes: Happy, Sad, Angry or Others.\footnote{Competition website: \url{https://www.humanizing-ai.com/emocontext.html}} The eight models developed in an incremental fashion, and submitted in that exact order to the competition (finally reaching rank 29/165) are made available together with a corresponding description of each iteration via a public repository.\footnote{Github repository: \url{https://github.com/zhaopku/ds3-emoContext}} The test data, consisting of 5,509 items was published by the organizers of the competition after its termination. This represents a non-adaptive scenario, where the developer does not get any direct feedback whilst submitting new models.

\begin{figure}
\centering
\includegraphics[width=0.75\textwidth]{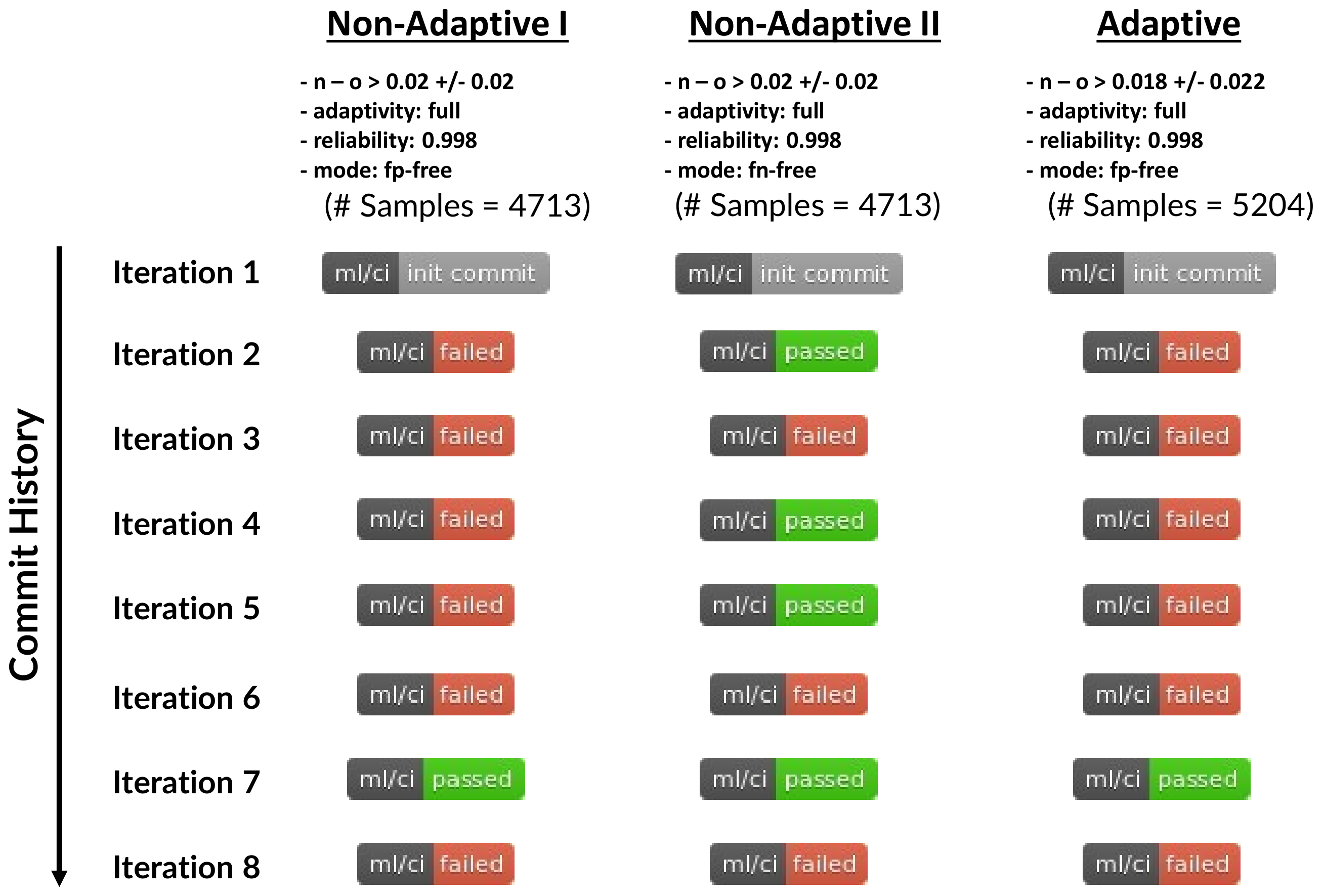}
\caption{Continuous Integration Steps in \sys.}
\label{fig:action}
\end{figure}

Figure~\ref{fig:action} illustrates three similar, but different test conditions, which are implemented in \sys.
The first two conditions check whether the new model is better than the old one by at least 2 percentage points in a non-adaptive matter. The developer will therefore not get any direct feedback as it was the case during the competition. While query (I) does reject false positive, condition (II) does accept false negative. The third condition mimics the scenario where the user would get feedback after every commit without any false negative.
All three queries were optimized by \sys using Pattern 2 and exploiting the fact that between any two submission there is no more than 10\% difference in prediction.

Simply using Hoeffding's inequality 
does not lead to a practical solution ---
for $\epsilon = 0.02$ and $\delta = 0.002$,
in $H=7$ non-adaptive steps, one would need
\[
n > \frac{r_v^2 (\ln H - \ln \frac{\delta}{2})}{2 \epsilon^2} = 44,268
\]
samples. This number even grows to up to 58K in the fully adaptive case!

All the queries can be supported rigorously with the 5.5K test samples provided after the competition. The first two conditions can be answered within two percentage point error tolerance and 0.998 reliability. The full-adaptive query in the third scenario can only achieve a 2.2 percentage point error tolerance, as the number of labels needed would be more than 6K, with the same error tolerance as in the first two queries.

\begin{figure}
\centering
\includegraphics[width=0.75\textwidth]{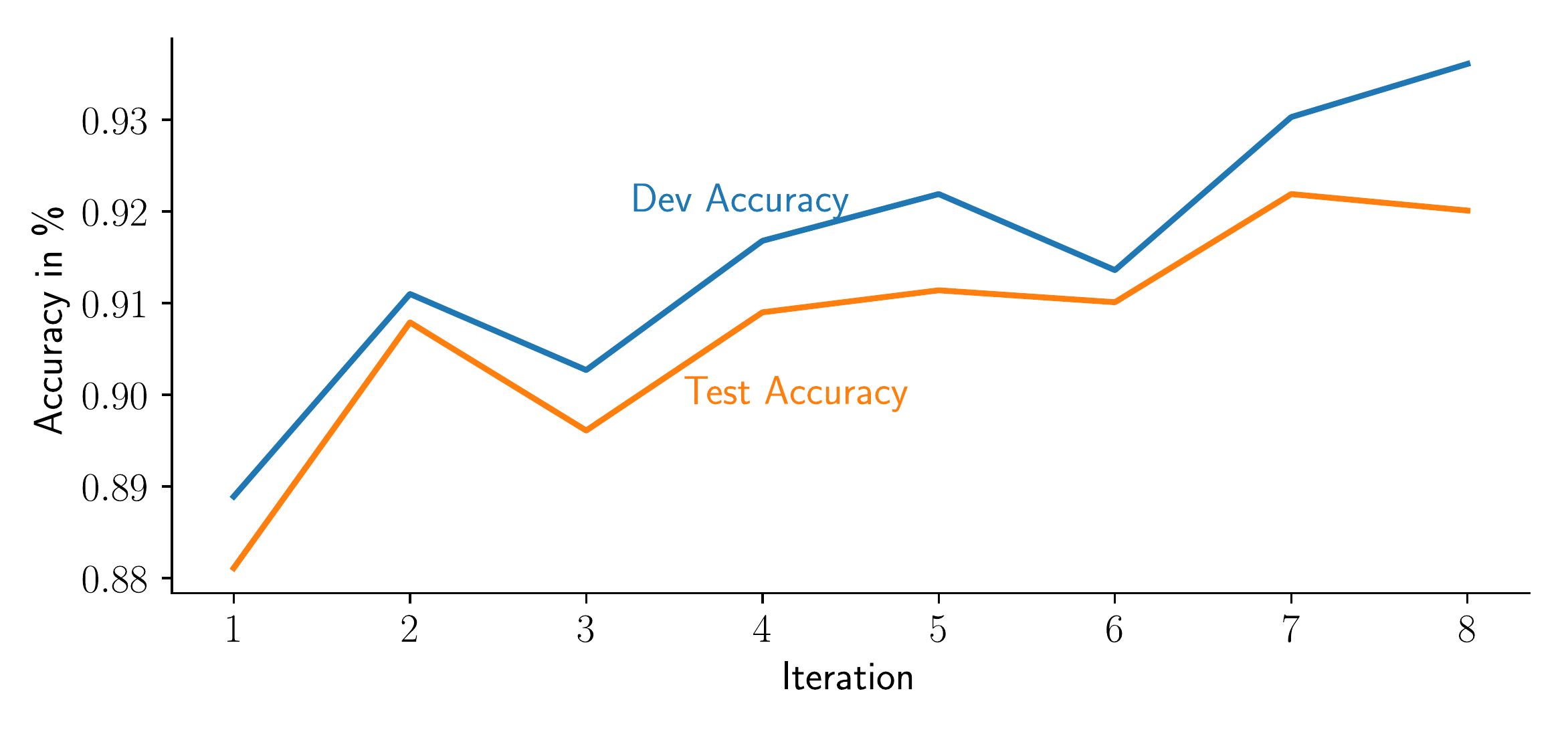}
\caption{Evolution of Development and Test Accuracy.}
\label{fig:trace_real}
\end{figure}

We see that, in all three scenarios,
\sys returns \texttt{pass}/\texttt{fail}
signals that 
make intuitive sense. If we look at the evolution of the development
and test accuracy over the eight iterations (see Figure~\ref{fig:trace_real}, the developer would ideally want \sys to accept her last commit, whereas all three queries will have the second last model chosen to be active, which correlates with the test accuracy evolution.

\section{Related Work}
\label{sec:relatedwork}

Continuous integration is a 
popular concept in software
engineering~\cite{duvall2007continuous}.
Nowadays, it is one of the best 
practices that most, if not all,
industrial development efforts follow.
The emerging requirement of a CI engine for
ML has been discussed informally in multiple
blog posts and forum discussions~\cite{url2, url5, url1, url3, url4}.
However, none of these discussions 
produce any rigorous solutions to testing the
quality of a machine learning model, which
arguably is the most important aspect of
a CI engine for ML. 
This paper is motivated by the 
success of CI in industry, and 
aims for building the first prototype
system for rigorous integration 
of machine learning models.

The baseline implementation of \sys builds on intensive
previous work on generalization
and adaptive analysis. The non-adaptive
version of the system is based on simple
concentration inequalities~\cite{boucheron2013concentration}
and the fully adaptive version of
the system is inspired by Ladder~\cite{blum2015ladder}. Comparing to the second, \sys is less restrictive on the feedback and more expressive given the specification of the test conditions. This leads to a higher number of test samples needed in general.
It is well-known that the $O(1/\epsilon^2)$
sample complexity of Hoeffding's inequality
becomes $O(1/\epsilon)$ when the
variance of the random variable $\sigma^2$ is 
of the same order of $\epsilon$~\cite{boucheron2013concentration}.
In this paper, we develop techniques to 
adapt the same observation to a real-world scenario (Pattern 1). The technique of
only labeling the difference between models
is inspired by disagreement-based active
learning~\cite{hanneke2014theory},
which illustrates the potential of taking
advantage of the overlapping structure
between models to decrease labeling 
complexity. In fact, the technique 
we develop implies that one can achieve
$O(1/\epsilon)$ label complexity
when the overlapping ratio between two
models $p = O(\sqrt{\epsilon})$.

The key difference between \sys and a differential privacy approach~\cite{dwork2014algorithmic}
for answering statistical queries lies in the
optimization techniques we design. By knowing
the structure of the queries we are able to considerably lower 
the number of samples needed.

Conceptually, this work is inspired
by the seminal series of work by Langford and others~\cite{langford2005tutorial, kaariainen2005comparison}
that illustrates the possibility for 
generalization bound to be practically tight.
The goal of this work is to build a practical
system to guide the user in employing 
complicated statistical inequalities and
techniques to achieve \emph{practical}
label complexity.

\section{Conclusion}
\label{sec:conclusion}

We have presented \sys, a continuous integration system for machine learning. It provides a declarative scripting language that allows users to state a rich class of test conditions with rigorous probabilistic guarantees. We have also studied the novel practicality problem in terms of labeling effort that is specific to testing machine learning models. Our techniques can reduce the amount of required testing samples by up to two orders of magnitude. We have validated the soundness of our techniques, and showcased their applications in real-world scenarios.

\subsection*{Acknowledgements}

{\small
We thank Zhao Meng and Nora Hollenstein for sharing their models for the SemEval'19 competition.
CZ and the DS3Lab gratefully acknowledge the support from Mercedes-Benz Research
\& Development North America, MeteoSwiss, Oracle Labs, Swiss Data Science Center, Swisscom, Zurich Insurance, Chinese Scholarship Council, and the Department of Computer Science at ETH Zurich.
}

\clearpage

\bibliographystyle{acm}
\bibliography{references}

\clearpage

\appendix

\section{Syntax and Semantics}
\label{app:syn_seman}

\subsection{Syntax of a Condition}
\label{app:syntax}
To specify the condition, which will be tested by
\sys whenever a new model is committed, the user
makes use of the following grammar:
{\small
\begin{verbatim}
      c   :- floating point constant
      v   :- n | o | d
      op1 :- + | -
      op2 :- *
      EXP :- v | v op1 EXP | EXP op2 c

      cmp :- > | < 
      C   :- EXP cmp c +/- c

      F   :- C | C /\ F
\end{verbatim}
}
\texttt{F} is the final condition, which is a 
conjunction of a set of clauses \texttt{C}.
Each clause is a comparison between an 
expression over $\{n, o, d\}$
and a constant, with an error tolerance 
following the symbol \texttt{+/-}.
For example, two expressions that we focus
on optimizing can be specified
as follows:
{\small
\begin{verbatim}
  n - o > 0.02 +/- 0.01 /\ d < 0.1 +/- 0.01
\end{verbatim}
}
in which the first clause
{\small
\begin{verbatim}
          n - o > 0.02 +/- 0.01
\end{verbatim}
}
requires that the new model have an accuracy that is
two points higher than the old model, with an error
tolerance of one point, whereas the clause
{\small
\begin{verbatim}
              d < 0.1 +/- 0.01
\end{verbatim}
}
requires that the new model can only change 10\%
of the old predictions, with an error tolerance of 1\%.

\subsection{Semantics of Continuous Integration Tests}
\label{app:semantics}

Unlike traditional continuous integration,
all three variables used in \sys, i.e., $\{n, o, d\}$,
are {\em random variables}. As a result, the 
evaluation of an \sys condition is inherently {\em probabilistic}. There are two additional parameters
that the user needs to provide, which would define 
the semantics of the test condition: (1) $\delta$,
the probability with which the test process is allowed 
to be incorrect, which is usually chosen to be smaller
than 0.001 or 0.0001 (i.e., 0.999 or 0.9999 success rate); and (2) \texttt{mode} chosen from
\texttt{\{fp-free, fn-free\}}, which specifies 
whether the test is {\em false-positive free} or
{\em false-negative free}.
The semantics are, with probability $1 - \delta$, the output of \sys is free of false positives or false negatives.

The notion of false positives or false negatives
is related to the fundamental trade-off between
the ``type I'' error and the ``type II'' error
in statistical hypothesis testing.
Consider
{\small
\begin{verbatim}
              x < 0.1 +/- 0.01.
\end{verbatim}
}

Suppose that the real {\em unknown} value of $x$ is $x^*$. Given an estimator $\hat{x}$, which,
with probability $1-\delta$, satisfies
\[\hat{x} \in [x^* - 0.01, x^* + 0.01], \]what should be the testing outcome of this condition? There are three cases:
\begin{enumerate}
\item When $\hat{x} > 0.11$, the condition should
return \texttt{False} because, given $x^* < 0.1$, the probability of having $\hat x > 0.11 > x^* + 0.01$ is less than $\delta$.
\item When $\hat{x} < 0.09$, the condition
should return \texttt{True} because, given $x^* > 0.1$, the probability of having $\hat x < 0.09 < x^* - 0.01$ is less than $\delta$.
\item When $0.09 < \hat{x} < 0.11$, the outcome cannot be determined: Even if
$\hat{x} > 0.1$, there is no way to tell
whether the real value $x^*$ is larger or smaller
than 0.1. In this case, the condition evaluates to
\texttt{Unknown}.
\end{enumerate}

The parameter \texttt{mode} allows the system to deal with the case that the condition evaluates to \texttt{Unknown}. In the \texttt{fp-free} mode, \sys treats \texttt{Unknown}
as \texttt{False} (thus rejects the commit) to ensure
that whenever the condition evaluates to \texttt{True}
using $\hat{x}$, the same condition is always \texttt{True} for $x^*$. 
Similarly, in the \texttt{fn-free} mode, 
\sys treats \texttt{Unknown}
as \texttt{True} (thus accepts the commit).
The false positive rate (resp. false negative rate)
in the \texttt{fn-free} (resp. \texttt{fp-free})
mode is specified by the error tolerance.

\end{document}